\documentclass[conference]{IEEEtran}
\IEEEoverridecommandlockouts
\usepackage{cite}
\usepackage{algorithm}
\usepackage{booktabs}
\usepackage{amsmath,amssymb,amsfonts}
\usepackage{algpseudocode}
\floatname{algorithm}{\textbf{Algorithm}}
\usepackage{graphicx}
\usepackage{textcomp}
\usepackage{multirow}
\usepackage{xcolor}
\usepackage{amsmath}
\usepackage{amssymb}
\usepackage{array}
\usepackage{float}
\usepackage{subcaption}

\usepackage{caption}
\usepackage{bm}
\usepackage{wrapfig}
\usepackage{graphicx}

\DeclareMathOperator*{\argmin}{argmin}
\author{\IEEEauthorblockN{Kawa Atapour\IEEEauthorrefmark{1},
		S. Jamal Seyedmohammadi\IEEEauthorrefmark{2},
		Jamshid Abouei\IEEEauthorrefmark{1},
		Arash Mohammadi\IEEEauthorrefmark{2},
		Konstantinos N. Plataniotis\IEEEauthorrefmark{3}}
	\IEEEauthorblockA{\IEEEauthorrefmark{1} Dept. of Electrical Engineering, Yazd University, Yazd, Iran}
	\IEEEauthorblockA{\IEEEauthorrefmark{2} Concordia Institute of Information Systems Engineering (CIISE), Concordia
		University, Montreal, Canada}
	\IEEEauthorblockA{\IEEEauthorrefmark{3} The Edward S. Rogers Sr. Dept. of Electrical \& Computer Engineering, University of Toronto, Canada}}

\def\BibTeX{{\rm B\kern-.05em{\sc i\kern-.025em b}\kern-.08em
    T\kern-.1667em\lower.7ex\hbox{E}\kern-.125emX}}

\begin{document}
\title{FedD2S: Personalized Data-Free Federated Knowledge Distillation\\
}
\maketitle

\begin{abstract}
This paper addresses the challenge of mitigating data heterogeneity among clients within a Federated Learning (FL) framework.
The model-drift issue, arising from the non-iid nature of client data, often results in suboptimal personalization of a global model compared to locally trained models for each client.
To tackle this challenge, we propose a novel approach named FedD2S for Personalized Federated Learning (pFL), leveraging knowledge distillation.
FedD2S incorporates a deep-to-shallow layer-dropping mechanism in the data-free knowledge distillation process to enhance local model personalization.
Through extensive simulations on diverse image datasets—FEMNIST, CIFAR10, CINIC0, and CIFAR100—we compare FedD2S with state-of-the-art FL baselines.
The proposed approach demonstrates superior performance, characterized by accelerated convergence and improved fairness among clients.
The introduced layer-dropping technique effectively captures personalized knowledge, resulting in enhanced performance compared to alternative FL models. 
Moreover, we investigate the impact of key hyperparameters, such as the participation ratio and layer-dropping rate, providing valuable insights into the optimal configuration for FedD2S. The findings demonstrate the efficacy of adaptive layer-dropping in the knowledge distillation process to achieve enhanced personalization and performance across diverse datasets and tasks.

\begin{IEEEkeywords}
	Personalized Federated Learning, Data-Free Knowledge Distillation.
\end{IEEEkeywords}
\end{abstract}

\section{\textbf{Introduction}}
\IEEEPARstart{D}{eep} Neural Networks (DNNs) have demonstrated remarkable performance across diverse artificial intelligence domains, such as computer vision, natural language processing, healthcare, and biometrics. Traditional learning paradigms involve a central entity collecting data from distributed sources to train DNNs. However, limitations in bandwidth for data transmission and privacy concerns of data sources impose constraints on the richness of collected data, hindering DNNs from reaching their full potential. Federated Learning (FL) emerges as a promising solution to address these challenges. By integrating FL with sensing and communication, a more robust and accurate global model can be developed, opening avenues for advanced artificial intelligence applications. Classical FL algorithms, exemplified by FedAvg \cite{FedAvg}, obtain the global model by iteratively averaging parameters of distributed local models, eliminating the need to access raw data. While FedAvg proves effective for real-world applications, its deployment poses practical challenges, including model-drift issues arising from non-iid data, model architecture and system heterogeneity \cite{FedAvg challenges, TPDS_DH, TPDS_SH}. System heterogeneity involves unbalanced communication, computation, and storage resources, particularly notable in mobile devices serving as clients. Under the non-iid setting, diverse class distributions, data size, and optimization objectives can cause weights to diverge, leading to a decline in global model performance. Additionally, the permutation-invariant property of neural networks poses challenges for element-wise aggregation methods in fully developing a global model.

In the context of FedAvg and the subsequent works, the process entails the training of a single global model. 
The challenge with this approach, however, is that heterogeneity in data distribution can decelerate the global model’s convergence, or even move it away from the global optima.
In such a situation, training local models exclusively on local datasets may produce better results compared to participating in an FL scenario.
To handle this situation, there has been a significant shift towards a new paradigm, named personalized Federated Learning (pFL) \cite{first_pFL}.
Invoking this new framework can lead to the advent of personalized models for every individual client, striking a balance wherein local knowledge is retained while still utilizing global information.
This enables clients to gain insights beyond their own training data and augment generalization.

Knowledge Distillation (KD) is a framework where the knowledge of a bulky, pre-trained model, named as the teacher model, is extracted and subsequently transferred to a lightweight, untrained student model. 
With regard to the fact that the main problem in pFL is how to exploit the knowledge from clients and how to transfer it to others, KD can be considered a potential approach to realize pFL \cite{pFedSD, TPDS_KD}.
This line of scientific investigation yields what is commonly referred to as Federated KD (FKD).
In this framework, the knowledge of the teacher model, corresponding to each input data, can be represented by the model’s response, intermediate features, or a relation of them.
The knowledge conveyed by responses refers to the output of the teacher model in the form of logits or soft labels.
Specifically, soft labels represent the relative probability of belonging each input data to classes and contain more didactic information about the input data in comparison to hard labels or ground truth.
This type of knowledge is often referred to as dark knowledge.
On the other hand, feature-based methods leverage knowledge extracted from intermediate layers of the teacher model in order to train the student model.
This newly incorporated knowledge serves as an additive term to its objective function, enabling the student model to emulate and replicate the behavior of the teacher model \cite{feature based improvement}.

\textbf{Related Works:} In the framework of FKD, clients and the server rely on a shared dataset, known as a public dataset, to synchronize the knowledge they generate.
The statistical properties of the public dataset play a key role in the effectiveness of the distillation process.
Essentially, this dataset must sufficiently represent the comprehensive data distribution spanned by all the clients to effectively mitigate the model-drift issue.
Nevertheless, the attainment of the public dataset by third-party institutions is impeded due to privacy considerations, rendering it impractical.
Consequently, the studies conducted by \cite{DFKD1}-\cite{DFKD4} are concerned with transferring acquired knowledge from the teacher model to the student model without the dependency on a public dataset.
In this regard, methods of Data-Free KD (DFKD) have been developed, in which the core idea is to generate synthetic data from a pre-trained model, as an alternative to the public dataset \cite{DFKD1}.
In many cases, an auxiliary model, typically a Generative Adversarial Network (GAN), is employed to generate synthetic data.

Building upon the DFKD paradigm, several initiatives have emerged to address the statistical heterogeneity of local datasets and the model architecture heterogeneity of clients, aiming to realize a Data-Free Federated Distillation (DFFD) framework within pFL.
Methods such as \cite{FedDTG, FedGEN, FedZKT} address the problem of dependency on the public dataset by training a GAN model and exchanging it between clients and the server.
Reference \cite{FedICT} proposes a mutual knowledge distillation framework \cite{mutual} to achieve DFFD, in which the client’s knowledge of the local data is transmitted to the server and distilled into a global model. Subsequently, the ensembled knowledge in the global model is transmitted back to the clients and distilled into respective local models.
Notably, in this paper, distillation occurs solely in the classifier part of local models.
In \cite{FedGEN}, clients train their personalized local models and then transmit the class distribution and the classifier part to the server.
The server aims to capture the knowledge of the global data distribution.
In this regard, it trains a GAN model to generate logits that closely resemble the logits obtained from an ensemble of the local classifier models.
Similar to the approach in \cite{FedICT}, the knowledge of the classifier part of local models is transmitted to the server as the learned knowledge of the clients.
However, according to \cite{Personalization_Layers, FedRep}, the personalized information of each client lies in the classifier part of the local model, and to capture personalization aspects of clients, the classifier part of models should not involved in the federated process, but rather be trained locally.

Motivated by the aforementioned observations and aiming to address the limitations of existing research, this paper introduces a new framework of DFFD for pFL, named Federated Deep-to-Shallow layer-dropping (FedD2S) . 
In FedD2S, we view each local model as a cascade of layers, in which deeper layers capture more personalized knowledge of the local dataset.
As the FL process progresses, we gradually restrict the involvement of deeper layers, and hence step by step, the personalized knowledge is maintained in the client.
Notably, this work pioneers the conceptualization of layers in the local model as distinct knowledge carriers, actively preventing the involvement of adverse knowledge in the federated learning process to enhance personalization. 
In addition, we apply this idea within a DFFD framework to accommodate practical constraints of data and model heterogeneity.

\textbf{Contributions:} Our main contributions can be summarized as follows:

\begin{enumerate}
	\item{We present FedD2S, an FD-based pFL framework that operates independently of public datasets. FedD2S enables personalized optimization on individual clients while mitigating the effects of client drift. This is achieved by gradually limiting the involvement of deeper layers' knowledge from other clients in the process of updating local models for each client.}
	
	\item{We extract the intermediate knowledge from local models and distill it into the global model using a head model constructed from dropped-layers of the global model. This method differs from the existing feature-based knowledge transfer methods in KD, such as \cite{FBDFKD1}, \cite{FBDFKD2}, and \cite{FBDFKD3}.}
	
	\item{We perform thorough experiments on FEMNIST, CIFAR10, CINIC10, and CIFAR100 datasets. Findings indicate that the proposed layer-dropping mechanism, as suggested in this study, enhances the average User model Accuracy (UA) across all compared baselines.}

\end{enumerate}

\section{\textbf{Preliminaries}}

\subsection{\textbf{Problem Statement}}
In this paper, we focus on a supervised $C$-class classification task in pFL.
The FL system consists of $N$ cooperative but heterogeneous clients, denoted by $\mathbb{U}=\{u_1, ...,u_N\}$, which are coordinated by a central server.
Each client $u_n\in\mathbb{U}$ possesses a local dataset, denoted by $\mathbb{D}^n = \bigcup_{i=1}^{K^n} \{(x_i^n, y_i^n)\}$, where $(x_i^n, y_i^n)$ represents $i^{th}$ data instance including the input and its ground-truth output, and $K^n$ denotes the size of the local dataset $\mathbb{D}^n$.
Each data instance $(x_i^n, y_i^n)$ is sampled from data distribution $\mathcal{D}^n$, where $\mathcal{D}^n$ is a distribution over data space $\mathcal{D}$, i.e., $\mathcal{D}^n\sim\mathcal{D}$.
In addition, we denote a batch of the input and output data by  $\boldsymbol X^n=[x_1^n, ..., x_{b}^n]$ and $\boldsymbol Y^n=[y_1^n, ..., y_{b}^n]$, respectively, where $b\leq K^n$ is the size batch.

Each client $u_n$ aims to train a local model parameterized by $\boldsymbol\theta^n=[\theta_{1}^n, ..., \theta_{L}^n]$, consisting of $L$ layers, where $\theta_{l}^n$ represents the parameter vector of $l^{th}$ layer.
In addition, the local model of client $u_n$ is dentoed by $F(\cdot; \boldsymbol\theta^n)$.
Throughout this paper, we assume that clients share the same network architecture.
Note that the intermediate output of the model in $l^{th}$ layer, per input data $\boldsymbol X^n$, is represented by $\boldsymbol H_l^n =[h_{l,1}^n,...,h_{l,b}^n]= F(\boldsymbol X^n;\boldsymbol{\theta}_{-l}^n)$, where $h_{l,i}^n$ is the intermediate output of $l^{th}$ layer per $i^{th}$ input data, and $\boldsymbol\theta_{-l}^n =[\theta_1^n, ..., \theta_{l}^n]$ indicates parameters of the model up to $l^{th}$ layer.

Conventional non-pFL methods aim to obtain a global model, parameterized by $\boldsymbol\theta^g$, which minimizes the total loss of clients across data space $\mathcal{D}$.
This is achieved through the following optimization problem \cite{FedAvg}, \cite{FedProx}:
\begin{equation}\label{FL_OP}
 		\boldsymbol{\theta}^{g,*} = \argmin_{\boldsymbol{\theta}^g}  \mathbb{E}_{\mathcal D^n\sim\mathcal{D}} \{ J(\boldsymbol{\theta}^g, \mathcal{D}^n)\},
\end{equation}
where $J(\boldsymbol\theta^g, \mathcal{D}^n)$ is the loss function of model $F(\cdot; \boldsymbol{\theta}^g)$ over $\mathcal{D}^n$, defined as follows:
\begin{equation}\label{loss function1}
 		J(\boldsymbol\theta^g, \mathcal{D}^n) = \mathbb{E}_{(x,y)\sim\mathcal{D}^n} \mathcal{L}_{CE} (F( x;\boldsymbol\theta^g), y),
\end{equation}
where $\mathcal{L}_{CE}$ denotes per-sample cross-entropy loss function.

In the pFL setting, due to the non-iid problem, optimizing a single global model does not generalize well on local datasets.
In such cases, training models locally may be more effective.
This may lead some clients to prefer training their local models on the local data and not participating in the FL process.
However, in practical situations, relying solely on the local dataset may not provide enough knowledge about the underlying task. 
To handle this contradiction, in the pFL setting, a personalized model $\boldsymbol{\theta}^n$ is considered for each client $u_n \in \mathbb{U}$ to match with its personalized characteristics.
In this regard, Eq. (\ref{FL_OP}) is converted to:

\begin{equation}\label{pFL}
 		\{\boldsymbol\theta^1, ..., \boldsymbol\theta^n\}^*= \argmin_{\{\boldsymbol\theta^1, ..., \boldsymbol\theta^n\}}\mathbb{E}_{\mathcal D^n\sim\mathcal{D}}\{ J(\boldsymbol{\theta}^n, \mathcal{D}^n)\}.
\end{equation}
Since the true distribution of local datasets are unknown, the local models are empirically optimized as follows:
\begin{equation}
 		\{\boldsymbol\theta^1, ..., \boldsymbol\theta^n\}^* = \argmin_{\{\boldsymbol\theta^1, ..., \boldsymbol\theta^n\}} \sum_{n = 1}^{N}\sum_{(x, y)\in \mathbb{D}^n}  \mathcal{L}_{CE} (F( x;\boldsymbol\theta^n), y).
\end{equation}
To solve this problem, each client needs to leverage the knowledge of other clients who are also working on the same task. 
In this regard, KD methods can be employed to create a framework that facilitates the sharing of knowledge. 
The details of KD are explained below.

\subsection{\textbf{Knowledge Distillation}}
KD is referred to any method of transferring knowledge from one or multiple teacher models into a student model \cite{KD}.
This process levearges a public dataset, denoted by $\mathcal D^p$, to align the mapping functions of teacher and student models. .
Specifically, the logit outputs of the teacher model, when applied to the public dataset, are passed through a softmax function to generate soft labels.
Along with the ground-truth outputs, which are used to train the student model in a conventional way, soft labels are utilized as regularizers to constrain the loss of the student model.
Typically, a Kullback-Leibler divergence function is employed to minimize the discrepancy between the soft labels of the teacher model and the predictions made by the student model, as follows \cite{KD}:
\begin{equation}
\begin{aligned}\label{KD}
	\boldsymbol\theta^{s,*} &= \argmin_{\boldsymbol\theta^s} \mathbb{E}_{(x,y)\sim \mathcal{D}^p} \bigg\{\mathcal{L}_{CE}(F^s( x; \boldsymbol\theta^s), y)\\
	&+ \tau^2 \mathcal{L}_{KL} (F^s( x; \boldsymbol\theta^s), F^t( x; \boldsymbol\theta^t)) \bigg\},
\end{aligned}
\end{equation}
where $F^s(\cdot; \boldsymbol\theta^s)$ and $F^t(\cdot; \boldsymbol\theta^t)$ are the student and teacher models, parameterized by $\boldsymbol\theta^s$ and $\boldsymbol\theta^t$, respectively. 
In addition, $\mathcal{L}_{KL}$ is the per-sample Kullback-Leibler loss function, and $\tau$ is the so-called temperature hyper-parameter used to soften generated logits.
Notably, throughout this paper, we assume that a soft-max function with temperature $\tau$ is employed in the output layer of models. 

In FKD, depending on the method, both clients and the server can play the role of either the teacher or the student.
In \cite{FedMD}, clients first update their local models using their respective local datasets, then, each client generates a set of soft labels by making predictions on a shared public dataset. On the server side, these local soft labels are averaged to create global soft labels, which represent global knowledge.
Finally, global knowledge is utilized in Eq. (\ref{KD}) to execute knowledge distillation.
This allows for the transfer of knowledge from other clients to each individual client.

The performance of knowledge sharing in FKD methods depends significantly on the distribution of the public dataset. However, in real-world scenarios, accessing a public dataset that accurately represents the entire distribution of local datasets is often impractical. Therefore, it is necessary to develop methods that facilitate the exchange of knowledge between clients and the server without the need for a shared dataset.
Most methods rely on generating synthetic data as a substitute for the public dataset and then employing traditional data-based KD techniques. These methods involve exchanging a generative model between clients and the server, resulting in a high communication overhead.
To address this issue, some studies aim to develop data-free methods that do not rely on generating synthetic data. Instead, these methods combine diverse local knowledge originating from heterogeneous datasets to obtain a comprehensive understanding of the task at hand.
The authors of \cite{FedICT} and \cite{FedGKT} utilize a global classifier in the server as a foundation for clients to transfer the local knowledge of local classifiers. However, these studies overlook the potential of transferring local knowledge presented within the feature extractor component of the models, which contains additional task-related information.
This paper presents a more generalized approach where a global model on the server acts as a repository, allowing clients to integrate various types of knowledge from their local models without using a public dataset.

\section{\textbf{Methodology: FedD2S Algorithm}}
In this section, we explain the FedD2S scheme in detail and present a summary version of it in Algorithm \ref{alg}.  
We also demonstrate the learning process through a visual representation in Fig. \ref{Fig1}.
\begin{center}
\vspace{-15pt}
\begin{figure*}[ht] 
       \includegraphics[scale=0.21]{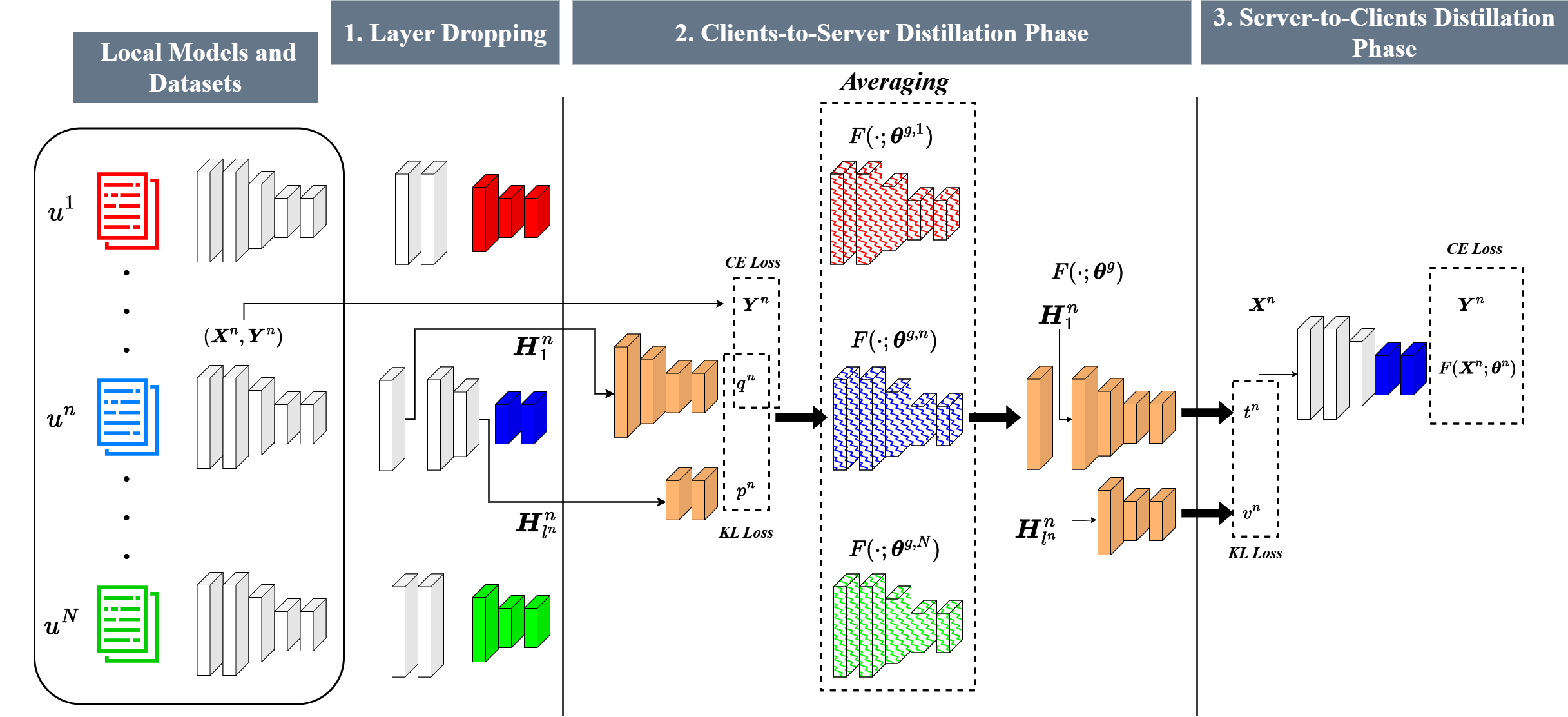}
	\caption{Illustration of the proposed FedD2S workflow.}
	\label{Fig1}
\end{figure*}
\end{center}
\vspace{-10pt}
The entire process of FL is executed within multiple rounds, denoted as $r \in \{1, ..., R\}$.
At the beginning of each communication round, a subset of clients, denoted by $\mathbb{U}^\rho$, where $\rho \in (0,1)$ represents the participation ratio, is activated to conduct local training on private datasets.
However, the local dataset is not rich enough to rely on, hence clients desire to join in an FL process and share their local knowledge.
Sharing knowledge between clients requires first ensembling local knowledge of diverse clients to obtain a global understanding, and then leverage that global understanding to enhance each client's performance.

KD methods provide an effective framework for ensembling knowledge and utilizing it to enhance the generalization of local models.
In this context, we adopt a mutual knowledge distillation approach \cite{mutual, FedICT}, which encompasses two phases: clients-to-server and server-to-clients knowledge distillation.
During the clients-to-server distillation phase, clients transfer their updated knowledge to the server.
On the server side, a global model $F(\cdot;\boldsymbol\theta^g)$ serves as a basis for clients to transfer their diverse local knowledge, resulting in a global view of the task.  Subsequently, the ensemble knowledge is transferred  back to individual clients and distilled into local models during the server-to-clients distillation phase.
The use of a centralized global model on the server facilitates the integration of diverse local knowledge from clients, eliminating the need for a public dataset.
Consequently, a knowledge-sharing framework is established in a data-free manner.

Typically, every KD method, including the two mentioned distillation phases, comprises two stages knowledge extraction and knowledge transferring.
Knowledge extraction refers to methods of extracting the captured knowledge of local models, typically in the form of tensors rather than model parameters.
This knowledge can be represented by the model's output, denoted by $\boldsymbol H_L = F(\boldsymbol X;\boldsymbol\theta)$, the intermediate layer's output, represented by $\boldsymbol H_l = F(\boldsymbol X;\boldsymbol\theta_{-l})$, or a combination of them.
Notably, the intermediate layer $\boldsymbol H_{l}$ can be either a feature map from a Convolutional Neural Network (CNN) layer or a vector from the output of a dense layer.
Knowledge transfer refers to any technique that enables the student model to reproduce the extracted knowledge of the teacher model.
In the subsequent sections, we will elaborate on these two stages and our main contribution in this regard.

\begin{algorithm}[t] 
	\renewcommand{\thealgorithm}{1}
	\caption{Proposed FedD2S}
	\label{alg}
	\begin{algorithmic}
		\State \textbf{Input:} local datasets $\{\mathbb D^n\}_{n=1}^N$
		\State \textbf{Initialization:} Learning rate $\alpha$, local models' parameters $\{\boldsymbol\theta_n\}_{n=1}^N$, number of layers $L$, participation ratio $\rho$.
		\For{$r = 1, \dots, R$}
		\State $\mathbb{U}^\rho \leftarrow$ a random selection of $\rho N$ clients from $\mathbb{U}$
		\For{all clients $u_n \in \mathbb U^\rho$ in parallel}
		\State $l^n =  \beta^n(r)$
		\EndFor
		\\\hrulefill
		\For{all clients $u_n \in \mathbb U^\rho$ in parallel}
		\For{all batches of $(\boldsymbol X^n, \boldsymbol Y^n) \subset \mathbb{D}^n$}
		\State $\boldsymbol H_{1}^n = F(\boldsymbol{X}^n;\boldsymbol\theta^n_{-1})$
		\State $\boldsymbol H_{l^n}^n = F(\boldsymbol{X}^n;\boldsymbol\theta^n_{-l^n})$
		\EndFor
		\State transmit all triplets $(\boldsymbol H_{1}^n, \boldsymbol H_{l^n}^n, \boldsymbol Y^n)$  to the server 
		\EndFor
		\For{all clients $u_n \in \mathbb U^\rho$}
		\State  $\boldsymbol \theta^{g,n} \leftarrow\boldsymbol\theta^{g}$
		\For{all triplets $(\boldsymbol H_1^n, \boldsymbol H_{l^n}^n, \boldsymbol Y^n)$}
		\State  $p^n = F(\boldsymbol H^n_{l_n}, \boldsymbol \theta^g_{+l^n})$
		\State  $q^n = F(\boldsymbol H^n_{1}; \boldsymbol \theta^g_{+1})$
		\State $\boldsymbol \theta^{g,n} \leftarrow\boldsymbol\theta^{g,n}- \alpha\nabla{J^n_{C2S}(\boldsymbol\theta^{g,n})}$
		\State $\boldsymbol \theta^{g,n} \leftarrow\boldsymbol\theta^{g,n}- \alpha\nabla{Q^n_{C2S}(\boldsymbol\theta^{g,n}))}$
		\EndFor
		\EndFor
		\State  $\boldsymbol \theta^{g} \leftarrow$ Average($\boldsymbol \theta^{g,n}, \forall u_n \in \mathbb{U}^\rho$)
		\\\hrulefill
		\For{$n\in\{1,...,N\}$ in the server}
		\State  $t^n =F(\boldsymbol H^n_{1},\boldsymbol \theta^g_{+1})$
		\State  $v^n =  F(\boldsymbol  H^n_{l^n}; \boldsymbol \theta^g_{ +l^n})$
		\State Transmit back $(t^n, v^n)$ to client $u_n$
		\EndFor
		\For{all clients $u_n \in \mathbb U^\rho$ in parallel}
		\State $\boldsymbol \theta^n\leftarrow\boldsymbol\theta^n- \alpha\nabla{J^n_{S2C}(\boldsymbol\theta^n)}$
		\State $\boldsymbol \theta^n\leftarrow\boldsymbol\theta^n- \alpha\nabla{Q^n_{S2C}(\boldsymbol\theta^n)}$
		\EndFor
		\EndFor
		\State \textbf{Output:} Personalized local models $\{F(\cdot;\boldsymbol\theta^n)\}_{n=1}^N$ 
	\end{algorithmic}
\end{algorithm}

Previous research \cite{Personalization_Layers}, \cite{FedRep} demonstrates that local models, consisting of both a feature extractor and a classifier, contain a higher degree of personalized knowledge within the classifier component than in the feature extractor part.
To capture the personalized aspects of the clients, the authors suggested locally updating the classifier without contributing to the FL process. 
Consequently, the draft issue arising from statistical heterogeneity is addressed, leading to an improvement in the clients’ performance on local test datasets.
Additionally, it is recognized that as layer depth increases in a CNN model, layers capture more abstract features \cite{CNN layers}, which represent more personalized characteristics of the dataset at hand.
This suggests that intermediate layers may contain different levels of personalized knowledge, and selecting which to incorporate into the FL process requires careful consideration.
Moreover, though the personalized knowledge of clients is heterogeneous, there is still a correlation among them.
This correlated information could be beneficial for other clients, suggesting that coordinating them across the FL process could potentially boost the performance.
Accordingly, in the preliminary stages of FL, before deep layers have fully captured personalized knowledge, this information could be shared among clients.
As the process progresses, this sharing can be curtailed to prevent over-assimilation into the FL process.
Building on these ideas, we propose a deep-to-shallow layer-dropping method (FedD2S), which generalizes algorithms presented in  \cite{Personalization_Layers}, \cite{FedRep}.

The proposed FedD2S approach preserves personalized layers both in the clients-to-server and server-to-clients distillation phases.
During the clients-to-server phase, personalized layers are excluded from participating in the federation and contributing to the ensemble knowledge.
This results in only partial knowledge of local models being transmitted to the server.
On the server side, the partial local knowledge is distilled to the total of global knowledge.
Henceforth, in this phase, the stream of knowledge is from partial local models to the total global model.
In the server-to-clients phase, personalized layers are not updated by the ensemble knowledge, hence, the ensemble knowledge from the total of global knowledge is transferred only to the partial local models.
As a result, the stream of knowledge is from the total global model to the partial of local models.

In the following sections, we offer a detailed explanation of the two phases of the mutual knowledge distillation approach and the stages of knowledge extraction and transferring involved in each phase.

\begin{figure*}[t]
	\centering
	\begin{subfigure}{0.32\textwidth}
		\includegraphics[width=\textwidth]{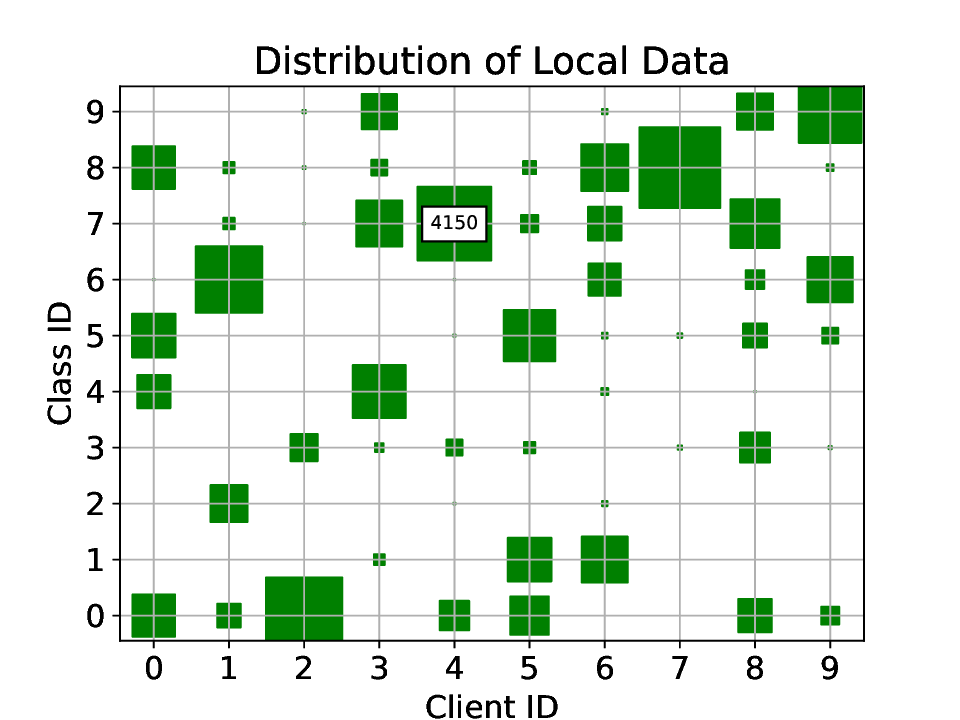}
		\caption{$\alpha=0.1$}
	\end{subfigure}
	\hfill
	\begin{subfigure}{0.32\textwidth}
		\includegraphics[width=\textwidth]{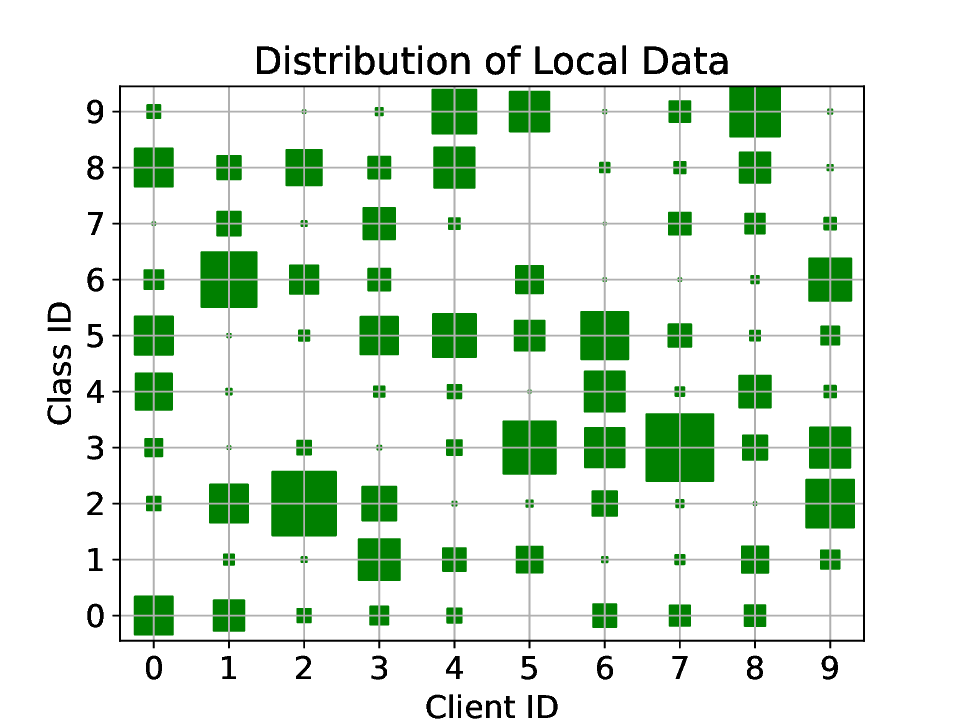}
		\caption{$\alpha=0.5$}
	\end{subfigure}
	\hfill
	\begin{subfigure}{0.32\textwidth}
		\includegraphics[width=\textwidth]{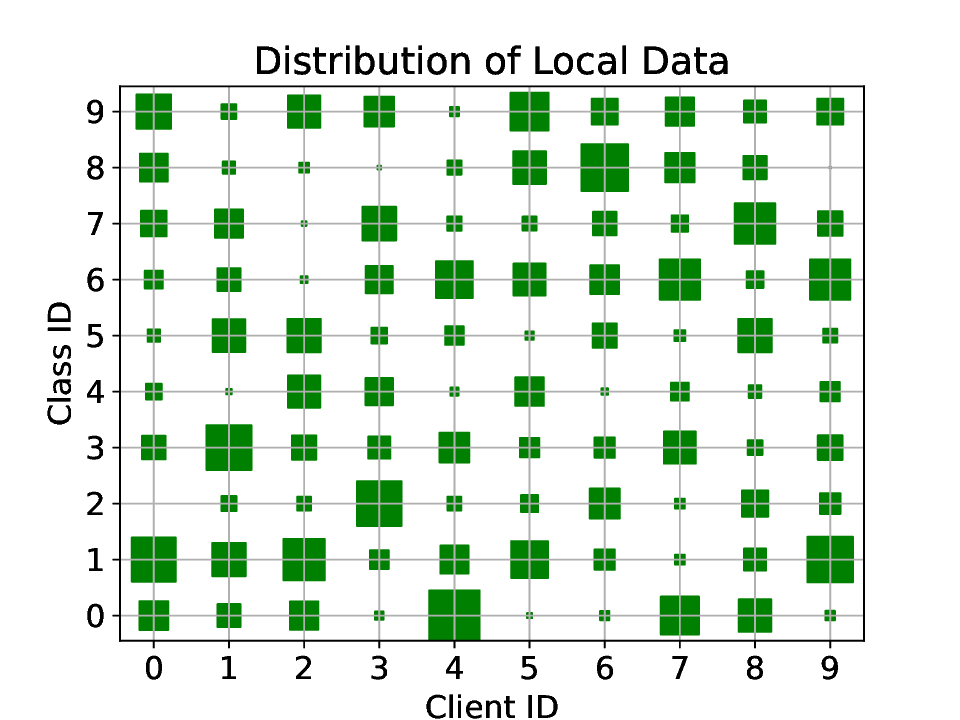}
		\caption{$\alpha=1$}
	\end{subfigure}
	\caption{Illustration of data heterogeneity among 10 clients on the CIFAR-10 dataset, where the x-axis shows client IDs, the y-axis indicates class IDs, and the size of squares indicates the number of training samples available for each class per client. For comparison, the number of samples for a square is reported in the left figure.} \label{fig_data_het}
\end{figure*}

\begin{table}
	\centering
	\caption{A summary of the four distinct soft labels outlined in this paper.}
	\begin{tabular}{|c|c|c|}
		\hline
		\textbf{Soft label} & \textbf{Generated by} & \textbf{Description} \\
		\hline
		$p^n$ & $F(\boldsymbol H^n_{l_n}, \boldsymbol \theta^g_{+l^n})$ & local knowledge before head is updated\\
		$q^n$ & $F(\boldsymbol H^n_{1}; \boldsymbol \theta^g_{+1})$ & ensemble knowledge before updating\\
		$v^n$ & $F(\boldsymbol  H^n_{l^n}; \boldsymbol \theta^g_{ +l^n})$ &  local knowledge after head is updated\\
		$t^n$ & $F(\boldsymbol H^n_{1},\boldsymbol \theta^g_{+1})$ & ensemble knowledge after updating\\
		\hline
	\end{tabular}
	\label{soft labels}
\end{table}

\subsection{\textbf{Clients-to-Server Distillation}}
\subsubsection{\textbf{Local Knowledge Extraction}}

In the proposed FedD2S algorithm, during the initial rounds of FL, the entire local model is involved in the federation process, and knowledge is extracted from the deepest layer, i.e., $\boldsymbol H_{L}$.
With progress, the deepest layer captures the personalized knowledge of the local dataset.
To mitigate the potential effects of this personalized knowledge on other local models, we drop this layer and prohibit it from participating in the federation process. 
From now on, the output of the next deepest layer, i.e., $\boldsymbol H_{L-1}$, represents the extracted knowledge and is comparatively less personalized than $\boldsymbol H_{L}$.
This process of layer-dropping continues, and in the last rounds of FL, only the concrete knowledge of clients, from shallower layers, is shared.
As a result, by progressively excluding the deeper layers in the FL process, personalization performance is enhanced.

The number of rounds required before dropping a layer varies depending on the level of personalization within the dataset and can differ across clients.
Consequently, layer-dropping is not universal, and the same layer in two different clients may present different levels of personalization.
The efficacy of FedD2S is highly influenced by the timing of layer dropout in local models.
Therefore, it is important to establish a strategy to determine the appropriate time for layer-dropping.
At this point, we define the function $l^n = \beta^n(\cdot)$, which determines the appropriate layer for knowledge extraction—referred to as the ``distillation layer``
We also introduce $Z_0$ as the duration, expressed in rounds, during which a layer is permitted to serve as the distillation layer, denoted as the ``dropping rate.`` 
Consequently, $\beta^n(\cdot)$ can be defined as follows:
\begin{equation}\label{dropping layer}
	\beta^n(r) =  L- \lfloor \frac{Z^n(r)-1}{Z_0} \rfloor, \forall u_n \in \mathbb{U},
\end{equation}
where $Z^n(r)$ denotes the number of rounds that client $u_n$ is selected to participate in the FL process by round $r$, and $\lfloor\cdot\rfloor$ denotes the floor function.
Since more levels of data heterogeneity result in more personalized layers, the rate of dropping layers needs to be a function of $\alpha$, hence we adopt different dropping rate $Z_0$ for different values of $\alpha$.


During each round, every client $u_n$ determines its distillation layer and extracts local knowledge $\boldsymbol H^n_{l^n}=F(\boldsymbol  X^n;\boldsymbol \theta^{n}_{-l^n})$.
The FedD2S algorithm enables clients to acquire local knowledge using their datasets instead of relying on a public dataset.
Consequently, each client’s local knowledge originates from a distinct dataset. 
Therefore, along with local knowledge of clients, the local dataset of each client is required to be sent to the server to train the global model.
However, sharing local datasets raises privacy concerns for clients.
To circumvent this, synthetic data can be generated using a GAN model and exchanged between clients and the server. 
Another approach involves sharing averaged representations of local data, known as “prototypes” \cite{Fed Prototype1}, \cite{Fed Prototype2}.
Alternatively, instead of sharing the local dataset, the output of the first layer, i.e., $\boldsymbol H^n_{1}=F(\boldsymbol X^n;\boldsymbol \theta^{n}_{-1})$ is transmitted to the server.
For simplicity, we adopt the latter approach by utilizing the output of the first layer.

\begin{center}
	\begin{table*}[ht]\centering 
		\caption{Average UA (\%) given different data settings, on various datasets FEMNIST, CINIC10, CIFAR10, and CIFAR100 with participation rate $\rho=0.2$.}\label{main_acc}
		\hspace*{-3.2em} 
		\begin{tabular}[t]{ m{1cm} m{1cm} m{1.5cm} m{1.5cm} m{1.5cm} m{1.5cm} m{1.5cm} m{1.5cm} m{1.5cm} m{1.5cm} m{1.7cm}}
			\hline
			Dataset&  IID Level&   FedAvg&  FedMD&   FedPer&   FedRep&  FedICT&    pFedSD&   FedPer+&   FedRep+&   FedD2S \\ 
			\hline
			FEMNIST&      $\alpha=0.1$\newline$\alpha=0.5$\newline$\alpha=1$&       $36.53\pm0.32$\newline$40.52\pm0.61$\newline$45.87\pm0.49$&           $49.27\pm0.46$\newline$51.65\pm0.92$\newline$58.46\pm0.53$& $51.33\pm0.56$\newline$54.09\pm0.77$\newline$59.48\pm0.69$&              $52.97\pm0.64$\newline$54.91\pm0.59$\newline$60.33\pm0.81$&           $53.35\pm0.43$\newline$56.01\pm0.39$\newline$64.96\pm0.41$& $54.44\pm0.87$\newline$54.99\pm0.77$\newline$65.90\pm0.33$&              $51.34\pm0.35$\newline$52.19\pm0.39$\newline$63.15\pm0.48$&           $53.04\pm0.86$\newline$51.85\pm0.55$\newline$63.96\pm0.68$&  $\bm{62.45\pm0.59}$\newline$\bm{57.11\pm0.31}$\newline$\bm{68.87\pm0.93}$\\
			\hline
			CIFAR10&       $\alpha=0.1$\newline$\alpha=0.5$\newline$\alpha=1$&       $59.91\pm2.16$\newline$63.24\pm0.89$\newline$65.10\pm0.23$&           $69.68\pm0.92$\newline$68.91\pm0.87$\newline$67.19\pm0.37$&            $70.93\pm0.45$\newline$68.21\pm1.09$\newline$67.37\pm0.32$&              $70.36\pm0.56$\newline$69.32\pm0.10$\newline$67.93\pm0.91$&           $76.47\pm0.37$\newline$72.15\pm0.18$\newline$72.91\pm1.04$&         $76.97\pm0.92$\newline$73.05\pm0.27$\newline$71.98\pm0.93$&              $74.92\pm0.23$\newline$73.08\pm0.35$\newline$72.01\pm0.38$&           $75.63\pm0.47$\newline$73.68\pm0.61$\newline$73.46\pm0.18$&           $\bm{77.47\pm0.29}$\newline$\bm{74.95\pm0.72}$\newline$\bm{74.01\pm0.84}$\\
			\hline 
			CINIC10&      $\alpha=0.1$\newline$\alpha=0.5$\newline$\alpha=1$&         $47.36\pm1.56$\newline$48.94\pm1.47$\newline$49.05\pm1.01$&           $57.41\pm1.30$\newline$58.48\pm0.12$\newline$64.17\pm0.55$&            $67.37\pm0.38$\newline$64.46\pm0.99$\newline$62.34\pm0.98$&              $68.35\pm0.56$\newline$65.56\pm0.89$\newline$64.75\pm0.99$&           $72.31\pm0.97$\newline$69.30\pm0.89$\newline$67.10\pm1.20$&         $74.16\pm0.81$\newline$70.71\pm1.07$\newline$67.88\pm0.79$&              $70.11\pm0.89$\newline$69.42\pm0.15$\newline$69.61\pm0.88$&           $72.64\pm0.75$\newline$71.31\pm0.56$\newline$71.06\pm0.71$&           $\bm{75.07\pm0.91}$\newline$\bm{72.15\pm0.32}$\newline$\bm{71.99\pm0.30}$\\
			\hline
			CIFAR100&    $\alpha=0.1$\newline$\alpha=0.5$\newline$\alpha=1$&        $28.11\pm0.48$\newline$30.47\pm0.22$\newline$35.39\pm0.16$&           $40.35\pm0.29$\newline$38.97\pm0.73$\newline$34.27\pm0.43$& $41.05\pm0.47$\newline$40.31\pm0.32$\newline$34.07\pm0.26$&              $41.91\pm0.21$\newline$41.02\pm0.35$\newline$36.01\pm0.96$&           $44.01\pm0.35$\newline$43.71\pm0.39$\newline$38.00\pm0.44$& $45.21\pm0.35$\newline$42.31\pm0.76$\newline$37.39\pm0.61$&              $44.01\pm0.27$\newline$41.03\pm0.19$\newline$37.00\pm0.18$&           $44.91\pm0.30$\newline$42.43\pm0.51$\newline$39.01\pm0.32$&      $\bm{47.41\pm0.75}$\newline$\bm{45.04\pm0.24}$\newline$\bm{38.91\pm0.37}$\\ 
			\hline
		\end{tabular}
	\end{table*}
\end{center}

\subsubsection{\textbf{Local Knowledge Transferring}}
Once local knowledge has been extracted from individual models (teacher models), the objective is to transfer this knowledge to the global model (student model).
The fundamental concept in this process is to enable the global model to replicate the local knowledge represented in $\boldsymbol H^n_{l^n}$.

Generally transferring intermediate knowledge extracted from the $l^{th}$ layer of the teacher model to the $l'^{th}$ layer of the student model, can be realized the following loss for each batch of data with size $b$ as follows: 
\begin{equation}
\begin{aligned}\label{loss function}
	J(\boldsymbol\theta^s, \mathcal{D}^p) = & \frac{1}{b}\sum_{i=1}^{b}  \{Dist( M^t(F(x,\boldsymbol\theta^t_{-l}))\\
										    & M^s(F(x, \boldsymbol\theta^s_{-l'}))\},
\end{aligned}
\end{equation}
where $M^t(\cdot)$ and $M^s(\cdot)$ are used to transform the intermediate outputs $F(x,\boldsymbol\theta^t_{-l})$ and $F(x,\boldsymbol\theta^s_{-l'})$ to the same dimensions, respectively.
Function $Dist(\cdot)$ refers to a loss function employed to reduce the discrepancy between two intermediate knowledge representations.
In \cite{FBDFKD1}, the $Dist(\cdot)$ function is defined as the Mean Squared Error (MSE) loss function, $M^s(\cdot)$ is implemented as a CNN network, and $M^t(\cdot)$ is an identical function.
On the other hand, in \cite{FBDFKD3}, the objective is to minimize the mutual information, leading to the utilization of negative log-likelihood as the $Dist(\cdot)$ function.

In this paper, for transferring the extracted knowledge of client $u_n$, i.e., $\boldsymbol H^n_{l^n}$, to the global model, a straightforward approach is to minimize the MSE between $\boldsymbol H^n_{l_n}$ and the corresponding intermediate knowledge of the global model, i.e., $\boldsymbol H^g_{l_n}$.
However, when the intermediate knowledge takes the form of a feature map, MSE may not serve as an appropriate distance criterion.
In this regard, we adopt a new approach, where the front part of the global model, which aligns with the rear part of the local model, is employed to map intermedite features into soft labels.
We call the front part of the global model the head model, denoted by $F(\cdot, \boldsymbol \theta^g_{+l})$, with $l$ representing any intermediate layer.
Specifically, the head model is utilized to transform intermediate feature $\boldsymbol H^n_{l^n}$ into soft labels $p^n$, as follows:
\begin{equation}\label{soft_labe1l}
	p^n = F(\boldsymbol H^n_{l_n}, \boldsymbol \theta^g_{+l^n}), \forall u_n \in \mathbb{U}, 
\end{equation}
where $\boldsymbol \theta^g_{+l^n} = [\theta_{l^n},...,\theta_{L^g}]$.
Next, the soft labels of the global model per $\boldsymbol H^n_{1}$ are extracted by $q^n = F(\boldsymbol H^n_{1}; \boldsymbol \theta^g_{+1})$.
To measure the discrepancy between soft labels $p^n$ and $q^n$, a Kullback-Leibler divergence function $\mathcal{L}_{KL}$ can be employed.
Therefore, Eq. (\ref{loss function}) is rewritten by substituting $M^t(\cdot)$, $M^s(\cdot)$, and $Dist(\cdot)$, with $F(\cdot, \boldsymbol \theta^{g}_{+l^n})$, the identical function,  and $\mathcal{L}_{KL}$, respectively:
\begin{equation}
 \begin{aligned}\label{C2S1} 
	 J^n_{C2S}(\boldsymbol \theta^{g}) = & \frac{1}{b}\sum_{i=1}^{b} \mathcal{L}_{KL}\biggl( F(h^n_{1,i};\boldsymbol \theta^{g}_{+1}) , \\
         &F( F(x^n_i; \boldsymbol \theta^n_{-l^n}); \boldsymbol \theta^{g}_{+l^n})\biggl ),  \forall u_n \in \mathbb{U}.
\end{aligned}
\end{equation}
After updating the global model by minimizing this loss function for each input data batch, we proceed to train the global model with ground-truth outputs by minimizing the following loss function:
\begin{equation}
 \begin{aligned}\label{C2S2} 
	 Q^n_{C2S}(\boldsymbol \theta^{g}) =   \frac{1}{b}\sum_{i=1}^{b} \mathcal{L}_{CE} \biggl(F(h^n_{1,i};\boldsymbol \theta^{g}_{+1}), y^n_i\biggl).
\end{aligned}
\end{equation}
By minimizing these two loss functions for each batch of input data, the local knowledge-transferring stage is accomplished \cite{FBDFKD1}.

The parameter vector of the updated global model for each client $u_n$ is then stored as $\boldsymbol \theta^{g,n}$.
Subsequently, all the updated global models are averaged to generate the final global model as follows:
\begin{equation}\label{soft_label2}
	\boldsymbol \theta^g = \frac{1}{N} \sum_{n=1}^{N} \boldsymbol \theta^{g,n}.
\end{equation}

\subsection{\textbf{Server-to-Client Distillation}}
To provide clients with a global perspective, a distillation phase where the ensemble knowledge of the global model is transmitted back to the clients is needed.
In the following, we elaborate on the global knowledge extraction from the global model and ensemble knowledge transferring into local models with more details.

\subsubsection{ \textbf{Global Knowledge Extraction}}
The ensembled knowledge of the global model per data $\boldsymbol H^n_{1}$ can be represented by the output of the global model in the form of soft labels, represented by $t^n = F(\boldsymbol H^n_{1},\boldsymbol \theta^g_{+1})$.
The stream of knowledge in this stage is from the total of the global model to the partial of the local model.
Specifically, ensemble knowledge $t^n$ is required to be transferred into only partial local model $F(\cdot;\boldsymbol \theta^n_{-l^n})$.
In this regard, the intermediate feature $\boldsymbol H^n_{l^n}$ is mapped by head model $F(\cdot; \boldsymbol \theta^g_{ +l^n})$, leading to soft labels $v^n = F(\boldsymbol  H^n_{l^n}; \boldsymbol \theta^g_{ +l^n})$.
Table \ref{soft labels} provides a consolidated overview of all four distinct soft labels described in this paper, presenting them together in a single location for enhanced clarity and reference.

\subsubsection{\textbf{Global Knowledge Transferring}}
Now, we instruct the client to replicate the global knowledge $t^n$. 
Subsequently, the loss function within this distillation phase is expressed as follows:
\begin{equation}
 \begin{aligned}\label{S2C1}
	J^n_{S2C}(\boldsymbol \theta^n_{-l^n}) = &\frac{1}{b}\sum_{i=1}^{b}  \mathcal{L}_{KL}\biggl( F( F(x^n_i; \boldsymbol \theta^n_{-l^n});\boldsymbol \theta^g_{+l^n}), \\	
				 &F(h^n_{1,i};\boldsymbol \theta^g_{+1}) \biggl ), \forall u_n \in \mathbb{U}.
\end{aligned}
 \end{equation}
Once the ensemble knowledge of the global model is transferred to the partial of the local model, the model is updated with ground-truth outputs as follows:
 \begin{align}\label{S2C2} 
	 Q^n_{C2S}(\boldsymbol \theta^{n}) =   \frac{1}{b}\sum_{i=1}^{b} \mathcal{L}_{CE} \biggl(F(x^n_i;\boldsymbol \theta^{n}), y^n_i\biggl).
\end{align}
To complete the global knowledge transfer stage, these two loss functions are minimized for each batch of data sequentially \cite{FBDFKD1}.


\section{\textbf{Simulation Results}}
Within this section, we concentrate on the performance overview and sensitivity analysis of the proposed method compared with other state-of-the-art baselines.
\subsection{\textbf{Simulation Setup}}
Unless stated otherwise, we conduct a total of $100$ communication rounds involving $50$ randomly selected clients, with a participation ratio of $\rho=0.2$.
Each communication round entails the random selection of participating clients.
We employ a local epoch $E=4$ with a batch size of $B=128$ samples during each distillation phase.
Average User model Accuracy (UA) serves as the benchmark metric for all methods, reflecting the average test accuracy of all local models.
The reported results represent the mean and standard deviation derived from three separate runs with distinct random seeds.

\textbf{Datasets:} We conduct simulations using four image datasets: FEMNIST \cite{FEMNIST}, CINIC10 \cite{CINIC10}, CIFAR10\cite{CIFAR10}, and CIFAR100 \cite{CIFAR10}. 
FEMNIST is an image dataset with 62 classes, covering 10 digits, 26 lowercase letters, and 26 uppercase letters.
CINIC10 and CIFAR10 are both 10-class classification datasets featuring everyday objects, while CIFAR100 is an extension of CIFAR10 with 100 classes.
The latter two datasets comprise 50,000 training samples and 10,000 testing samples each, whereas CINIC10 includes 90,000 training samples.
 Each classification task involves distributing the entire dataset among $N$ clients, where $K^n = \frac{|\mathbb{D}|}{N}$, and $|\mathbb{D}|$ represents the dataset's cardinality.
Additionally, each client allocates 20 percent of its local dataset for a dedicated local test set.

\textbf{Data Heterogeneity:} To account for the varying distribution of local data among clients, we use a Dirichlet distribution, as suggested by previous studies \cite{FedGEN}, \cite{FedICT}.
This continuous probability distribution enables us to model probabilities over a set of categories.
We represent Dirichlet distribution as $Dir(\alpha)$, where $a$ controls the degree of non-iid-ness.
Smaller values of $a$ result in more skewed and therefore more non-iid data.
Specifically, for each client $u_n\in\mathbb{U}$, we sample a vector $d^n\sim Dir(\alpha)$ with the size of a number of classes, where $j^{th}$ component of $d^n$ determines the number of samples from class $j$. 
We illustrate the impacts of different values of $\alpha$ on the distribution of the CIFAR10 dataset across 10 clients in Fig. \ref{fig_data_het}.

\textbf{Model Architecture:}
In our simulations, we utilize CNN architectures integrated with fully connected networks.
Specifically, we employ two different model architectures, denoted as $M1 = [C_1(8); C_2(16); C_3(32); F_1(32); F_2(16); F_3(10)]$ and  $M2 = [C_1(16); C_2(64); C_3(128); F_1(128); F_2(32); F_3(10)]$, where $C_i(j)$ represents the $i^{th}$ CNN layer with $j$ channels, and $F_i(j)$ signifies the $i^{th}$ dense layer with a size of $j$ neurons. 
It is worth noting that a flattened layer is incorporated between the CNN and dense layers.
We deploy model $M1$ for CINIC10 and CIFAR10 datasets, while $M2$ is employed for FEMNIST and CIFAR100 datasets.

\textbf{Baselines:} We compare the proposed FedD2S with several state-of-the-art methods, classified into four groups: $(1)$ non-pFL methods (FedAvg \cite{FedAvg}, FedMD \cite{FedMD}), $(2)$ pFL data-based methods (FedPer \cite{Personalization_Layers}, FedRep \cite{FedRep}), $(3)$ pFL data-free methods (FedICT \cite{FedICT}, pFedSD \cite{pFedSD}), and $(4)$ extensions of the previous methods incorporating a layer-dropping mechanism (FedPer+, FedRep+).

In the pioneer approach for FL, FedAvg, local model parameters are averaged to create a global model.
In the case of FedMD, after local updates, each client generates soft labels on a public dataset, transmitting them to the server.
The server then averages these local soft labels to create global soft labels, which are subsequently used to distill global knowledge from the server into each local model.

In FedPer, local neural networks are divided into two components: the base and personalization. 
The base is shared across all participants to create the global representation, while the personalization layers are exclusively updated during local training.
FedRep incorporates the concept of partitioning the local model into representation and head components. In this approach, the outputs from the representation part of various clients are transmitted to the server and averaged. The resulting averaged representation is then employed by clients to update their respective head components.

FedICT introduces a strategy of decoupling local models into feature extractor and classifier components. 
This approach employs a mutual knowledge distillation technique to facilitate knowledge exchange between local and global classifiers.
The process involves transmitting the locally extracted feature map, along with local logits, from the local feature extractor to the server. 
These data are utilized to train a global classifier, and the resulting logits are subsequently transmitted back to clients to distill the global knowledge into their respective local classifiers.
In pFedSD, the server initiates the process by broadcasting the global model to participating clients.
Each client initializes its local model with the received global model, followed by conducting local training with a self-knowledge distillation mechanism. 
Clients store the updated local model as the teacher for the subsequent round and transmit this updated model back to the server. 
Finally, the server aggregates all received local models to derive a new global model. This cooperative interaction ensures an iterative refinement of the global model through the collaboration of the server and participating clients.

FedPer+ and FedRepr+ represent extensions within our proposed configuration, departing from the conventional practice of extracting knowledge from a fixed intermediate layer. Instead, these extensions leverage the innovative dynamic mechanism of layer-dropping introduced in our approach.

\textbf{Hyperparameters:}
We employ the Adam optimizer with a fixed learning rate of 0.01 across all baseline models.
In our proposed FedD2S approach, we specifically the dropping layers set $C_3, [F_1, F_2, F_3]$. 
For FedPer and FedRep, we extract intermediate knowledge from the flattened layer situated between $C_3$ and $F_1$.
The selection of optimal dropping rates $Z_0$ is contingent upon the varying levels of data heterogeneity, leading to configure $Z_0$ as $3$, $5$, and $7$ for distinct values of $\alpha = 0.1, 0.5, 1$, respectively. 
In the interest of fairness, for all baselines incorporating a knowledge distillation stage, we conduct simulations with $\frac{E}{2}=2$ epochs dedicated to local training and an additional $\frac{E}{2}=2$ epochs for the distillation process.

\begin{figure*}[t]
	\vspace{-15pt}
	\centering
	\begin{subfigure}{0.32\textwidth}
		\includegraphics[width=\textwidth]{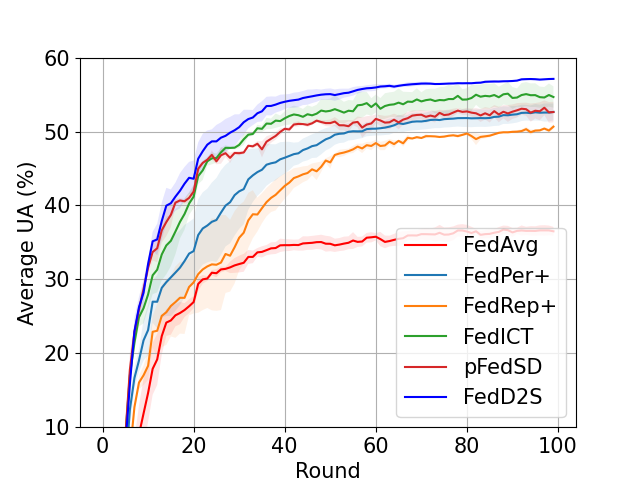}
		\caption{FEMNIST}
	\end{subfigure}
	\hspace{3cm}
	\begin{subfigure}{0.32\textwidth}
		\includegraphics[width=\textwidth]{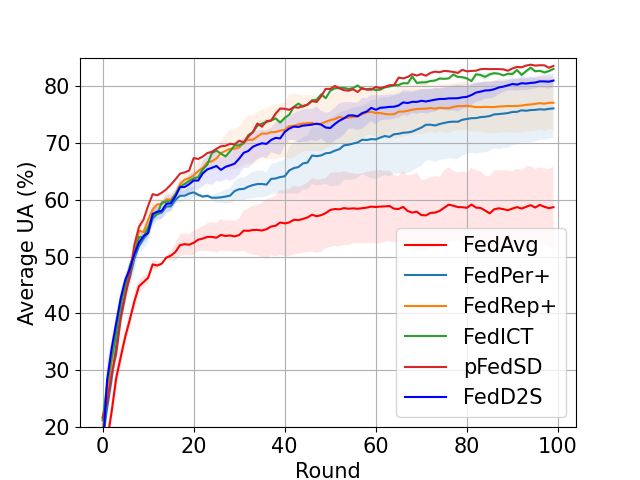}
		\caption{CIFAR10}
	\end{subfigure}
	\begin{subfigure}{0.32\textwidth}
		\includegraphics[width=\textwidth]{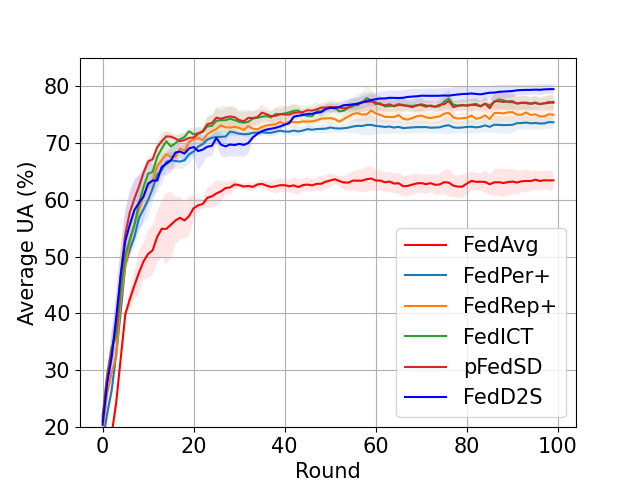}
		\caption{CINIC10}
	\end{subfigure}
	\hspace{3cm}
	\begin{subfigure}{0.32\textwidth}
		\includegraphics[width=\textwidth]{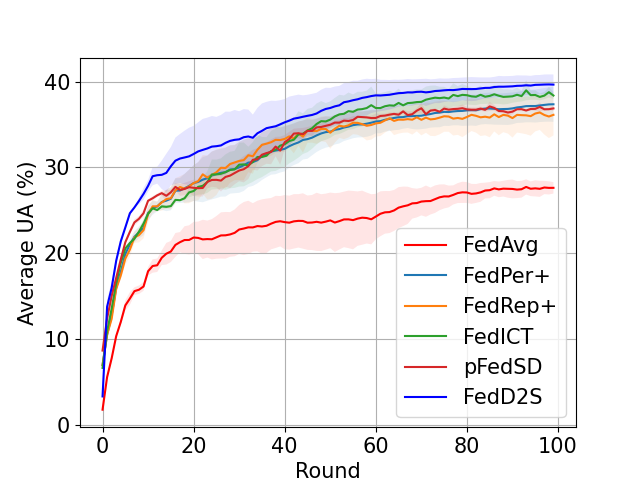}
		\caption{CIFAR100}
	\end{subfigure}
	\caption{Learning curves of average UA (\%) of the proposed FedD2S compared to baseline methods across different datasets, with $\rho=0.2$, $\alpha=0.1$, and $Z_0=3$.} \label{curves}
\end{figure*}

\subsection{\textbf{Simulation Results and Performance Analysis}}
\textbf{Accuracy:} Table \ref{main_acc} reports the average UA of local test datasets along with its standard variation across different levels of heterogeneity $\alpha=0.1, 0.5, 1$ for various baseline models.
The average UAs are computed by averaging the local accuracies of all clients over the last 10 communication rounds.
As indicated, the proposed model surpasses the performance of the best baseline. 
In our simulations, the results consistently demonstrate that both FedPer+ and FedRep+ baselines outperform FedPer and FedRep across all datasets.
This superior performance can be attributed to the incorporation of a layer-dropping mechanism in these methods.
This observation underscores the effectiveness of the layer-dropping mechanism as a key factor in improving the personalization performance of federated learning models.
Among the non-FedD2S baselines, both FedICT and pFedSD consistently exhibit superior performance. FedICT leverages personalized head models, contributing to its enhanced efficacy in capturing individual client characteristics and thereby achieving notable results. Similarly, pFedSD stands out due to its incorporation of a self-distillation mechanism, showcasing the effectiveness of knowledge distillation in improving model performance.

\textbf{Convergence:}
In this section, we analyze learning curves across different datasets and baselines to assess their convergence rate and determine the requisite number of communication rounds for achieving a specific accuracy.
The learning curves in Fig. \ref{curves} are generated under the conditions of $\alpha=0.1$, with a single epoch allocated for each local training or distillation process, involving 50 clients with $\rho=0.2$, and dropping rate $Z_0=3$.
The results demonstrate that the proposed FedD2S method showcases accelerated and smoother convergence, surpassing the performance of alternative baselines.
Additionally, it is worth mentioning that the proposed FedD2S achieves the same accuracy with the fewest communication rounds.

\begin{figure*}[htp]
 \vspace{-15pt}
  \centering
  \begin{subfigure}{0.33\textwidth}
    \includegraphics[width=\textwidth]{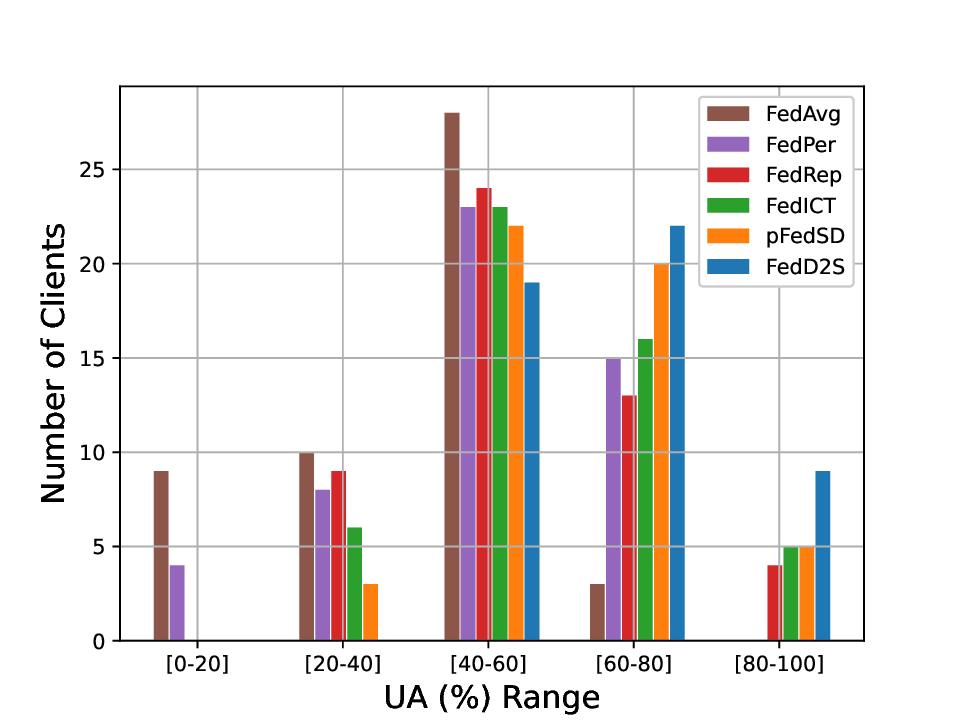}
    \caption{FEMNIST}
  \end{subfigure}
\hspace{3cm}
  \begin{subfigure}{0.33\textwidth}
    \includegraphics[width=\textwidth]{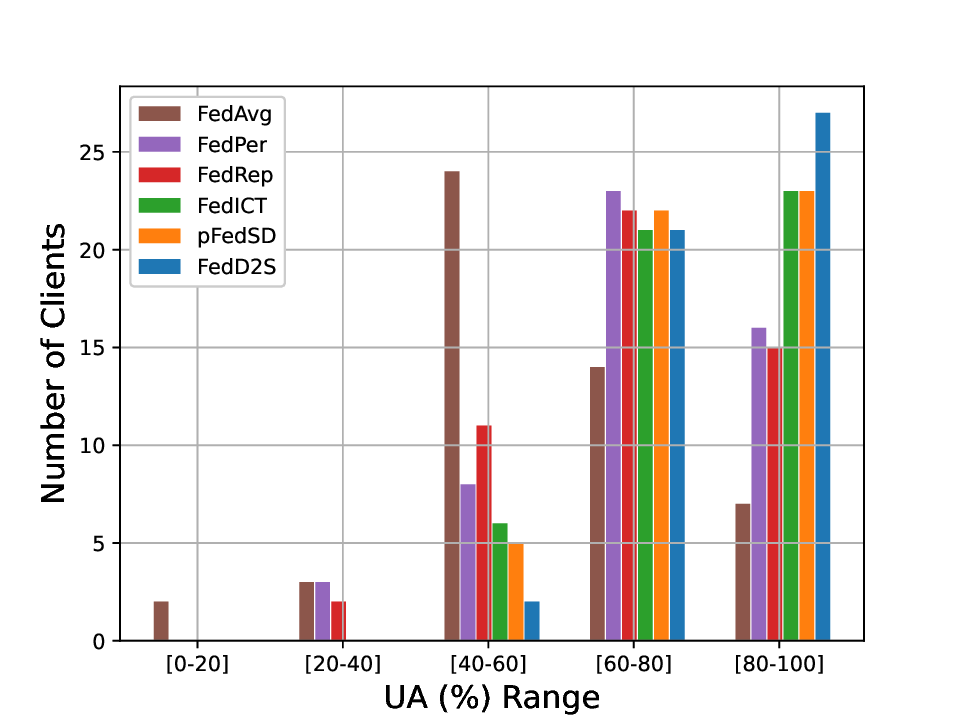}
    \caption{CIFAR10}
  \end{subfigure}
  \begin{subfigure}{0.33\textwidth}
    \includegraphics[width=\textwidth]{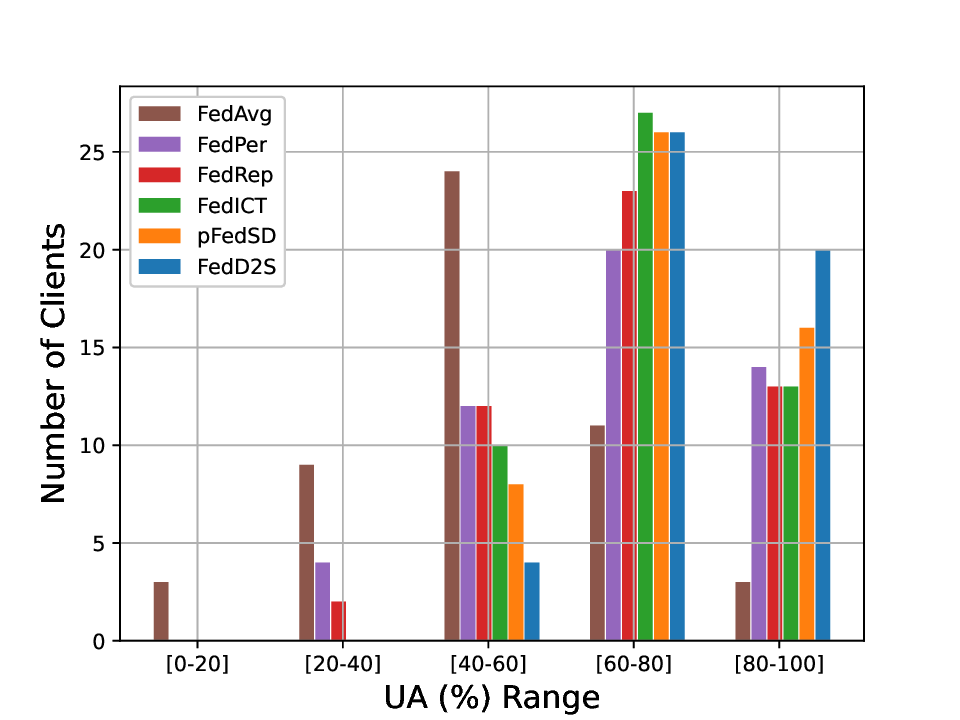}
    \caption{CINIC10}
  \end{subfigure}
\hspace{3cm}
  \begin{subfigure}{0.33\textwidth}
    \includegraphics[width=\textwidth]{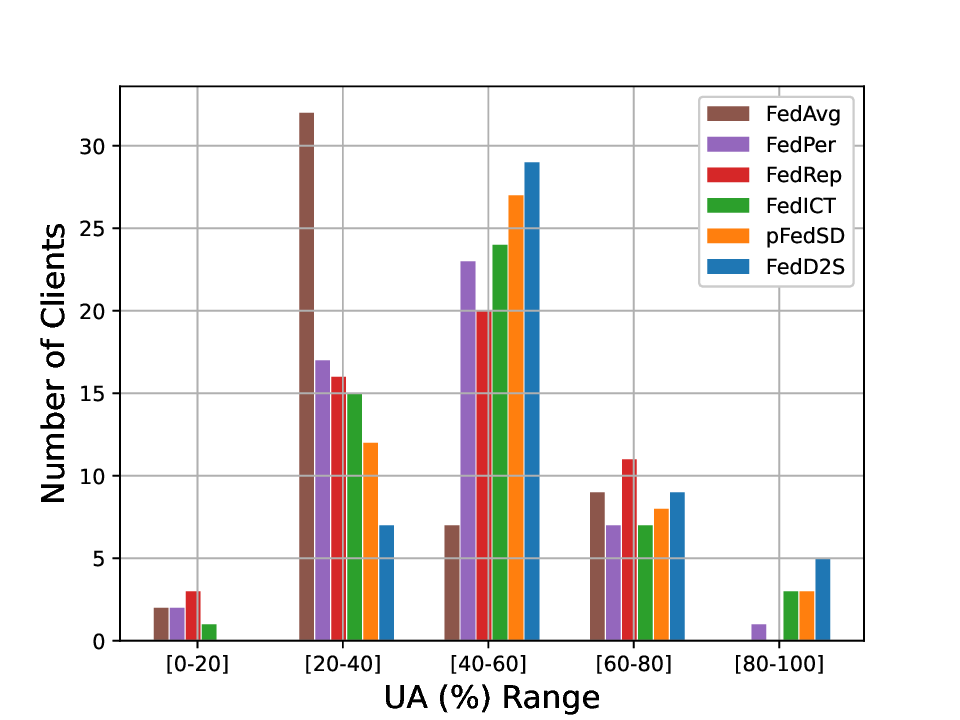}
    \caption{CIFAR100}
  \end{subfigure}
  \caption{Comparison of client distribution across accuracy ranges for different datasets—FEMNIST, CIFAR10, CINIC10, and CIFAR100—under the conditions $\alpha=0.1$ and $\rho=0.2$.} \label{Fairness}
\end{figure*}

\textbf{Fairness among clients:}
Given the variability in data characteristics, individual clients may present significant variations in their personalized performance. 
To assess the fairness of the enhancements in individualized performance, we scrutinize the effectiveness of the distinct models held by each client. 
Fig. \ref{Fairness} illustrates the overall spread of client performance, revealing that FedD2 consistently exhibits a higher count of clients attaining elevated testing accuracy across all datasets.

\textbf{The performance of KD with head-models:}
In this part, we evaluate the performance of the proposed FedD2S with a head-model configuration.
We compare its performance with a configuration where head-models are not utilized, but instead, the distance between the intermediate layers of the client and the server is reduced using the MSE loss function.
Table \ref{headmodel} presents a comparison across various layer-dropping sets. 
In this assessment, we explore four distinct dropping sets, denoted as $\mathcal{L}_1=[F1, F2, F3]$, $\mathcal{L}_2=[F2, F3]$, $\mathcal{L}_3=[C3, F1, F2, F3]$, and $\mathcal{L}_4=[C2, C3, F1, F2, F3]$. 
The simulations are conducted on the CIFAR-10 dataset. The results indicate that the proposed method with head-models exhibits superior performance compared to the configuration employing the MSE loss function. Notably, for $\mathcal{L}_3$ and $\mathcal{L}_4$ dropping sets, there is a significant performance gap between the two configurations. We posit that this discrepancy arises from the inadequacy of the MSE loss function in capturing the difference of intermediate knowledge in the form of feature maps.

\begin{table}[ht] \caption{Comparison of average UA (\%) for the proposed method with head-model configurations and MSE loss function across four different layer-dropping sets.}\label{headmodel}
\begin{tabular}[t]{ m{1.5cm} m{1.5cm} m{1.5cm} m{1.5cm}  m{1.5cm} }
  \hline
  Configuration&                         \multicolumn{1}{c}{$\mathcal{L}_1$}&           \multicolumn{1}{c}{$\mathcal{L}_2$}&      \multicolumn{1}{c}{$\mathcal{L}_3$}&    \multicolumn{1} {c}{$\mathcal{L}_4$}\\
  \hline
  FedD2S with head-models&       $78.23\pm0.72$&            $76.83\pm0.85$&         $77.16\pm0.73$&        $75.23\pm0.86$\\
  \hline
  FedD2S with MSE&                   $76.85\pm0.49$&            $75.46\pm0.31$&         $74.17\pm0.66$&        $53.57\pm0.78$\\ 
  \hline
\end{tabular}
\end{table}

\subsection{\textbf{Sensitivity Analysis}}
\begin{figure}[htp]
  \vspace{-15pt}
  \centering
  \begin{subfigure}{0.22\textwidth}
    \includegraphics[width=\textwidth]{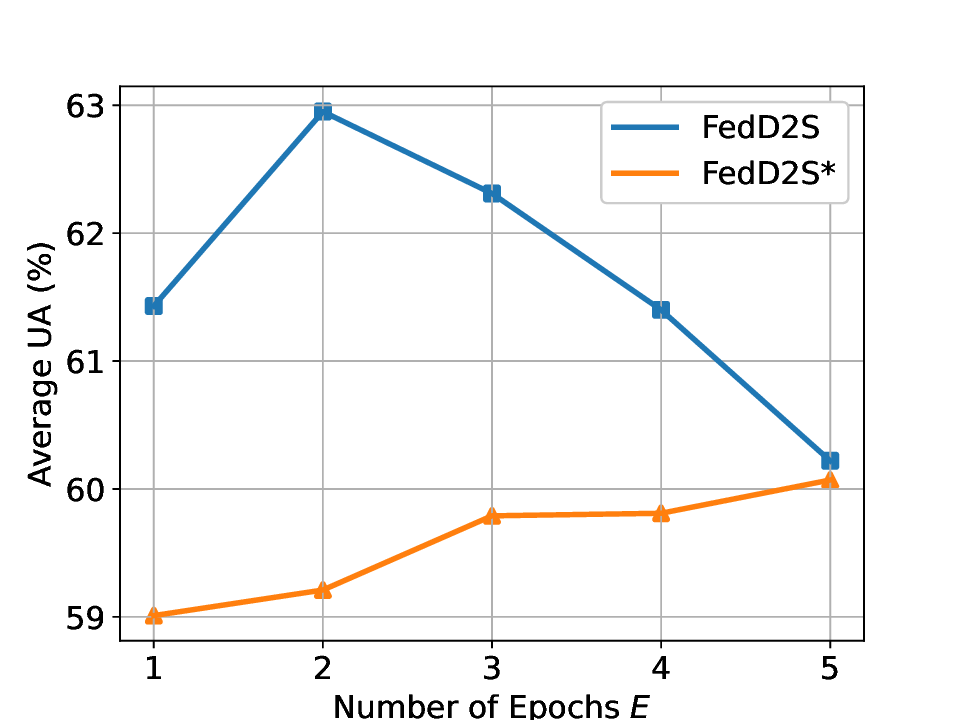}
    \caption{FEMNIST}
  \end{subfigure}
  \begin{subfigure}{0.22\textwidth}
    \includegraphics[width=\textwidth]{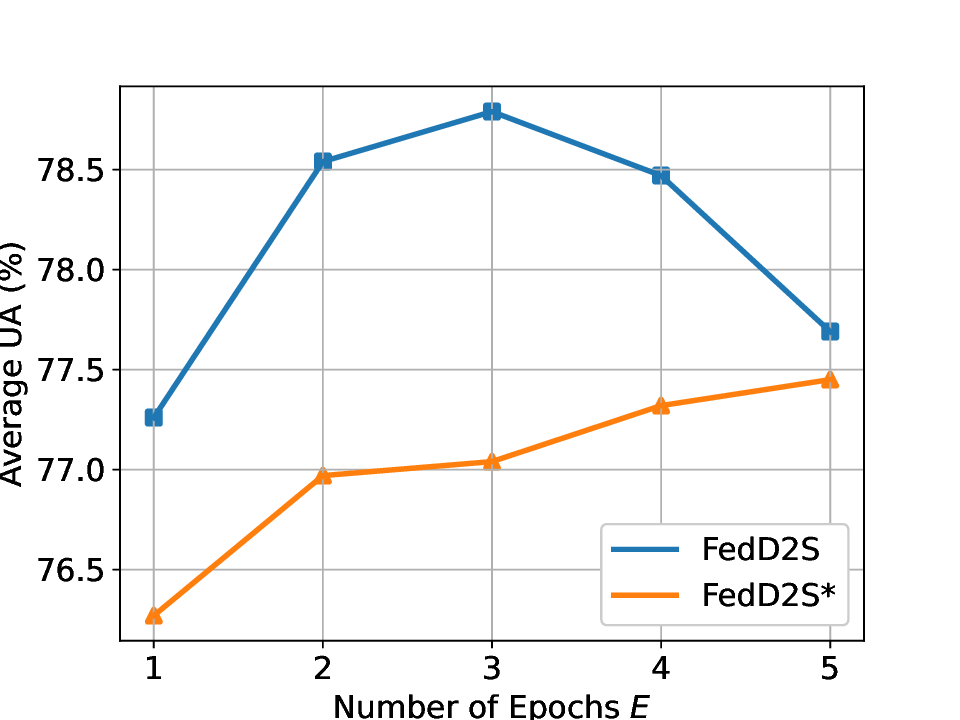}
    \caption{CIFAR10}
  \end{subfigure}
  \begin{subfigure}{0.22\textwidth}
    \includegraphics[width=\textwidth]{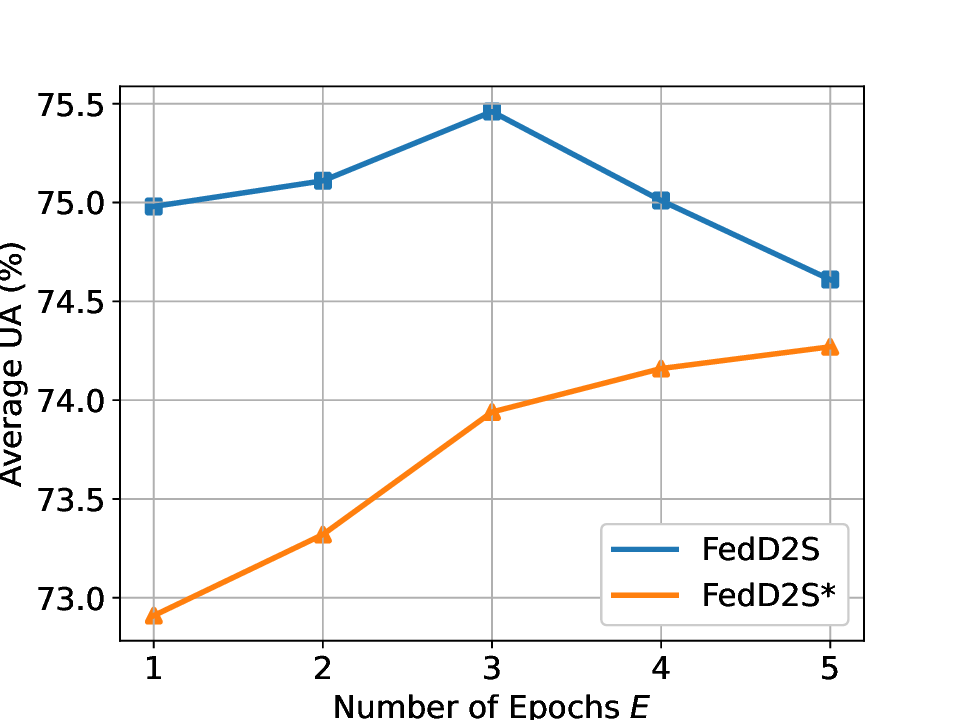}
    \caption{CINIC10}
  \end{subfigure}
  \begin{subfigure}{0.22\textwidth}
    \includegraphics[width=\textwidth]{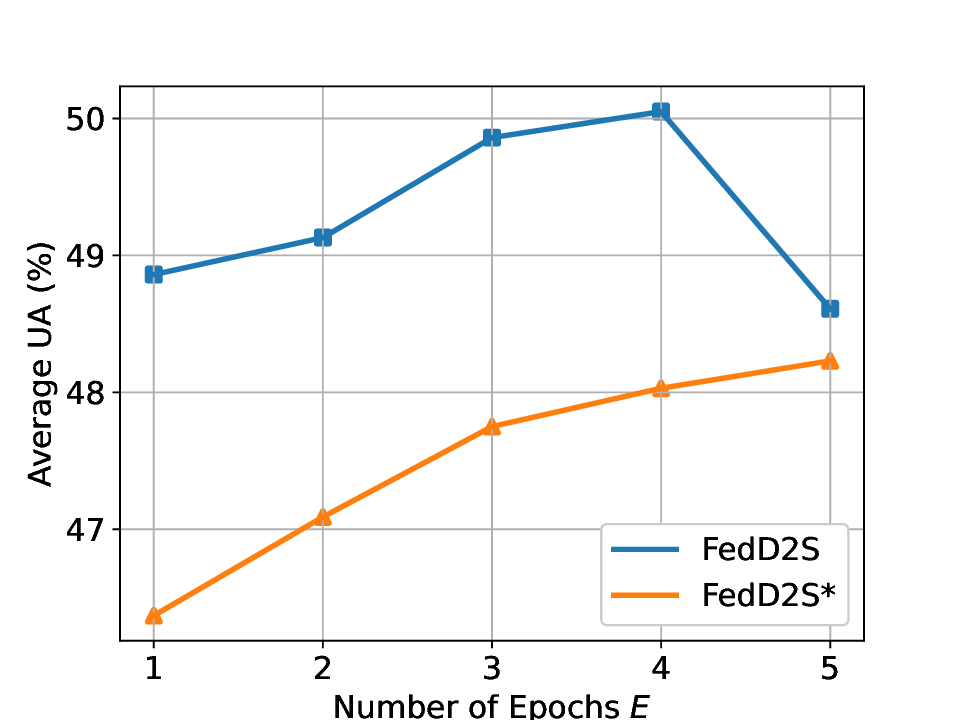}
    \caption{CIFAR100}
  \end{subfigure}
  \caption{The influence of layer-dropping dynamics with varying epochs across different datasets.} \label{epochs}
\end{figure}

\textbf{Effects of the number of epochs:}
The effectiveness of our proposed deep-to-shallow layer-dropping technique lies in initially incorporating deep layers into the FL process and subsequently dropping them before fully capturing the personalized knowledge from other clients.
Therefore, in the configuration of the proposed method, it is crucial to ensure that layers are dropped before getting too converged.
Fig. \ref{epochs} illustrates the performance of the proposed FedD2S method and a baseline, denoted as FedD2S*, where only the head part of the local models is shared.
Notably, the graph reveals that as the number of epochs increases, the performance of FedD2S initially improves but then declines, whereas the performance of FedD2S* consistently rises. 
This arises from the fact that higher numbers of epochs make deeper layers to capture the personalized knowledge of other clients, which consequently ruins the personalization.
This observation aligns with our assertion that layers should be dropped before they converge, representing a key limitation of our proposed method.

\textbf{Effects of different layer-dropping sets:}
Fig. \ref{dropping_sets} depicts the average UA curves against FL rounds for diverse layer-dropping sets applied to the CIFAR10 dataset.
The figure highlights that an optimal layer-dropping set should neither be too small nor too large; it should strike a balance to encompass both personalization and sharing aspects.
This is crucial, since During the last rounds of the FL process, the proposed model acts similarly to the FedPer baseline.

\begin{center}
\vspace{-15pt}
\begin{figure}[ht] 
       \includegraphics[scale=0.5]{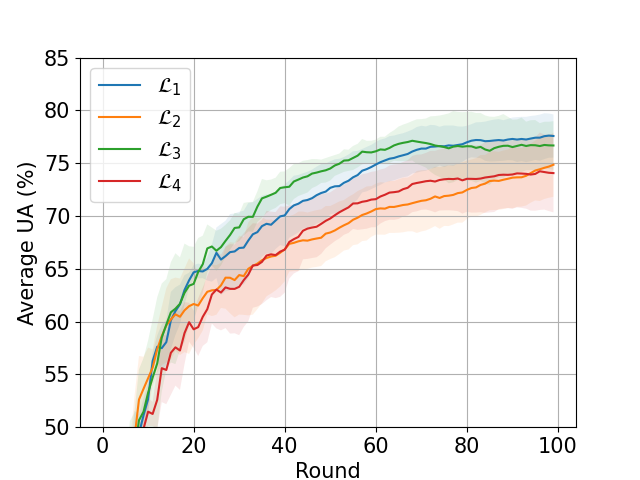}
\caption{Average UA (\%) curves across Federated Learning rounds for different layer-dropping sets on the CIFAR10 dataset, with $\alpha=0.1$ and $Z_0=3$.} \label{dropping_sets}
\end{figure}
\end{center}

\textbf{Effects of different values of layer-dropping rate:}
Fig. \ref{various distillation intervals} provides a visual representation of the diverse impact of the dropping rate $Z_0$ on the performance of the proposed FedD2S method. 
The experiments, conducted on the CIFAR100 dataset, incorporate vertical lines indicating the standard deviation across three distinct runs with different seeds.
The observed trend reveals that an initial increase in the dropping rate $Z_0$ leads to performance enhancement, followed by a subsequent decline.
The results highlight that augmenting heterogeneity among local datasets correlates with a decrease in the optimal average UA.
This phenomenon arises because higher heterogeneity levels prompt shallower layers to more rapidly capture personalized knowledge from clients, necessitating earlier layer-dropping.
Conversely, lower heterogeneity implies reduced personalization in local datasets, allowing layers to leverage knowledge from other clients for a more extended period.
Notably, higher heterogeneity corresponds to a higher deviation from the mean value. 
This visualization reinforces our approach in formulating Eq. (\ref{dropping layer}) as a function of $\alpha$ demonstrating the intricate interplay between dropping rate, performance trends, and dataset heterogeneity within the proposed FedD2S method.

\section{\textbf{Conclusion}}
The proposed FedD2S federated learning approach presented a promising paradigm for overcoming challenges associated with distributed training across heterogeneous client datasets.
By introducing a dynamic deep-to-shallow layer-dropping mechanism, FedD2S demonstrated superior performance in terms of User model Accuracy compared to state-of-the-art personalized federated learning baselines.
The comprehensive simulation results, spanning diverse datasets and experimental configurations, underscored the method's effectiveness in achieving higher average local accuracy, accelerated convergence, and ensuring fairness among participating clients.
The study shed light on the nuanced interplay of dropping rates, epochs, and layer configurations, offering valuable insights into the potential and limitations of the proposed FL approach. 
FedD2S contributed to the ongoing discourse on efficient and collaborative federated learning strategies, paving the way for further advancements in addressing real-world challenges associated with decentralized model training.

\begin{center}
	\vspace{-15pt}
	\begin{figure}[ht] 
		\includegraphics[scale=0.5]{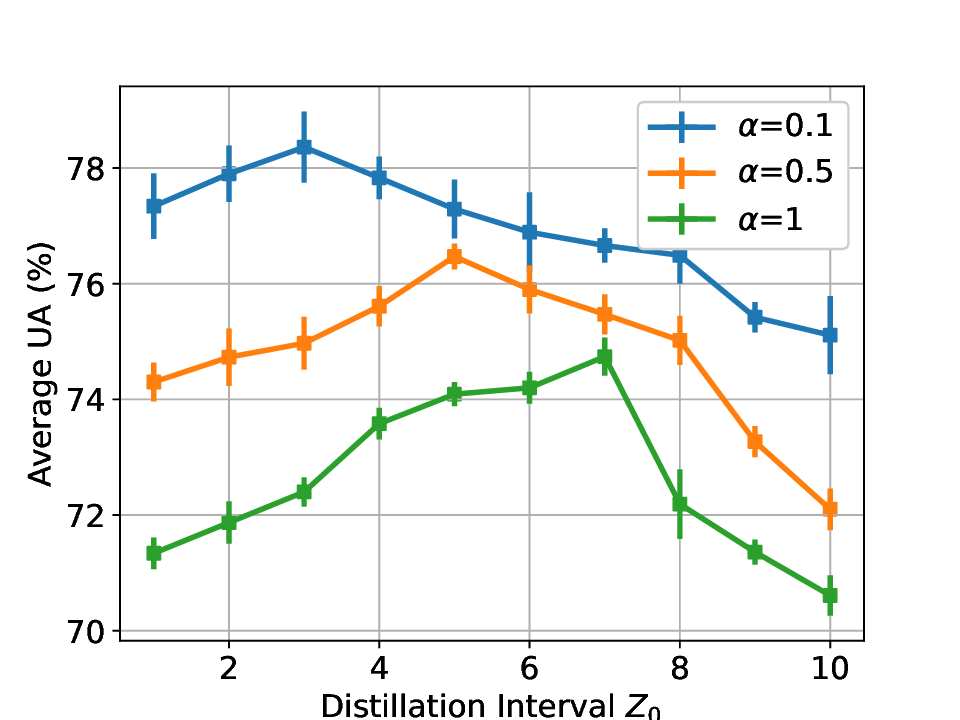}
		\caption{Average UA (\%) on FedD2S performance with varying data heterogeneity $\alpha$ for different dropping rate $Z_0$ on CIFAR100.} \label{various distillation intervals}
	\end{figure}
\end{center}

\vspace{-38pt}
\begin{IEEEbiography}[{\includegraphics[width=1in,height=1.25in,clip,keepaspectratio]{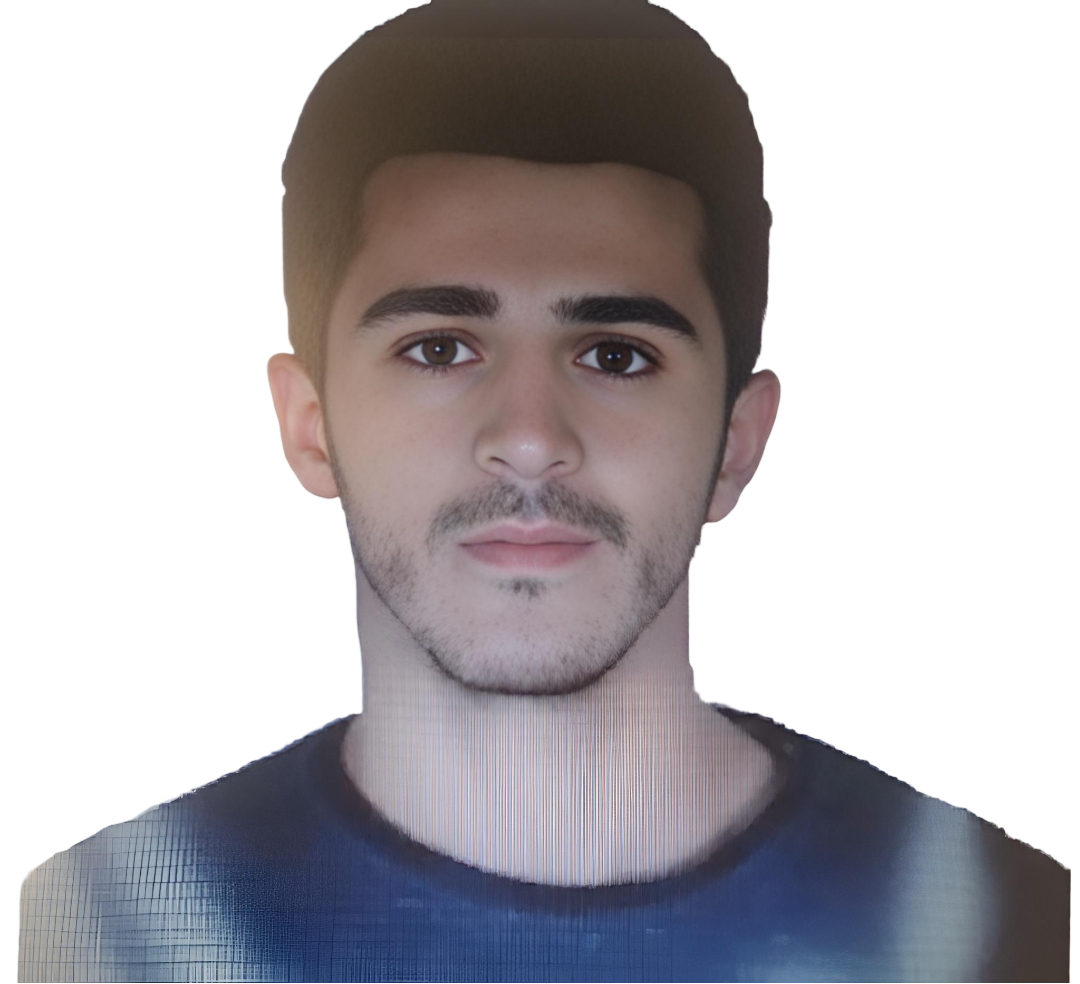}}]{S. Kawa Atapour}
Obtained a Bachelor of Science degree in Electrical Engineering from Bu-Ali Sina University, Iran, and later achieved a Master of Science degree in Communication Systems Engineering from Tarbiat Modares University, Iran, in 2017 and 2021, respectively. His research focuses on signal processing, machine learning, federated learning, and reinforcement learning.
\end{IEEEbiography}
\vspace{-50pt}
\begin{IEEEbiography}[{\includegraphics[width=1in,height=1.25in,clip,keepaspectratio]{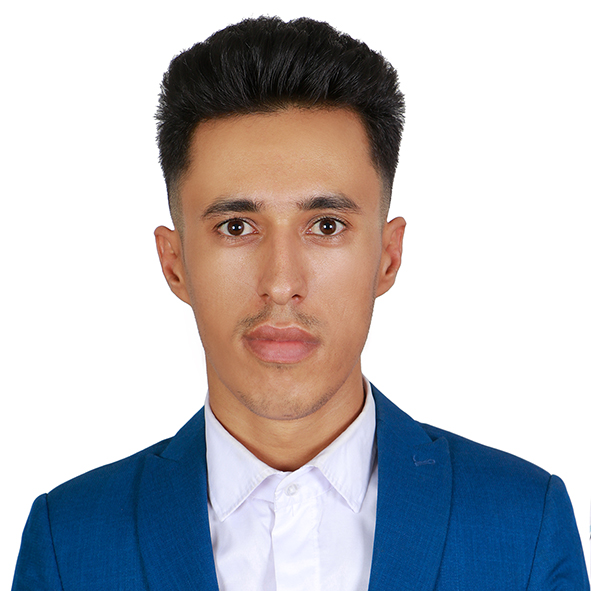}}]{S. Jamal Seyedmohammadi}
	received his B.Sc. degree in Power Engineering from the
	Bu-Ali Sina University, Iran, and M.Sc. degree
	in communication-Systems Engineering from
	Iran University of Science and Technology, Iran,
	in 2017 and 2021, respectively. He is currently a PhD candidate at Concordia University. His research
	interests include signal processing, machine
	learning, and federated learning.
\end{IEEEbiography}
\vspace{-40pt}
\begin{IEEEbiography}[{\includegraphics[width=1in,height=1.25in,clip,keepaspectratio]{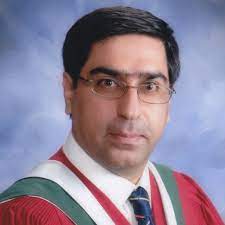}}]{Jamshid Abouei}
	received the B.Sc. degree in Electronics Engineering and
	the M.Sc. degree in Communication Systems
	Engineering both from
	Isfahan University of Technology (IUT), Iran,
	in 1993 and 1996, respectively, and the Ph.D.
	degree in Electrical Engineering from University
	of Waterloo, Canada, in 2009. He is currently a Professor with
	the Department of Electrical Engineering, Yazd
	University, Iran. He focuses his research primarily on the following areas: 5G/6G Wireless networks, mobile edge caching, federated learning, and hybrid RF/VLC. From
	2009 to 2010, he was a Postdoctoral Fellow in the Multimedia Lab,
	in the Department of Electrical and Computer Engineering, University of
	Toronto, Canada, and worked as a Research Fellow at the Self-Powered
	Sensor Networks (ORF-SPSN) consortium. During his sabbatical, he was
	an Associate Researcher in the Department of Electrical, Computer and
	Biomedical Engineering, Ryerson University, Toronto, Canada. Dr Abouei
	was the International Relations Chair in 27th ICEE2019 Conference, Iran,
	in 2019. Currently, Dr Abouei directs the research group at the Wireless
	Networking Laboratory (WINEL), Yazd University, Iran. His research
	interests are in the next generation of wireless networks (5G) and wireless
	sensor networks (WSNs), with a particular emphasis on PHY/MAC layer
	designs including the energy efficiency and optimal resource allocation in
	cognitive cell-free massive MIMO networks, multi-user information theory,
	mobile edge computing and femtocaching. Dr Abouei is a Senior IEEE
	member and a member of the IEEE Information Theory.

\end{IEEEbiography}
\vspace{-40pt}
\begin{IEEEbiography}[{\includegraphics[width=1in,height=1.25in,clip,keepaspectratio]{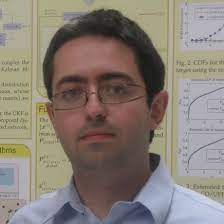}}]{Arash Mohammadi}
	received the B.Sc. degree from the ECE Department, University of Tehran, Tehran, Iran, in 2005, the M.Sc. degree from the BME Department, Amirkabir University of Technology (Tehran Polytechnic), Tehran, in 2007, and the Ph.D. degree from the EECS Department, York University, in 2013. From 2013 to 2015, he was a Postdoctoral Fellow with the Multimedia Laboratory, ECE Department, University of Toronto. He is currently an Associate Professor with the Concordia Institute for Information Systems Engineering (CIISE), Concordia University, Montreal, QC, Canada. His reseach interests include machine learning, biomedical signal/image processing, statistical signal processing. He was the Director of Membership Developments of IEEE Signal Processing Society (2018–2021), and the General Co-Chair of 2021 IEEE International Conference on Autonomous Systems (ICAS). Additionally, he was a member of the Organizing Committee of 2023 IEEE Intelligent Vehicles Symposium (IV 2023), the 2021 IEEE International Conference on Acoustics, Speech and Signal Processing (ICASSP), and the 2021 IEEE International Conference on Image Processing (ICIP). He is also the Program Chair of the 2024 IEEE International Conference on Human-Machine Systems (IEEE ICHMS) and is on the editorial board of IEEE Signal Processing Letters and Scientific Reports (Nature). 
\end{IEEEbiography}
\vspace{-40pt}
\begin{IEEEbiography}[{\includegraphics[width=1in,height=1.25in,clip,keepaspectratio]{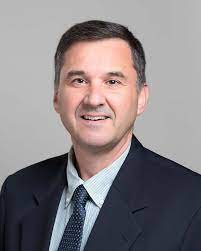}}]{Konstantinos N. Plataniotis}
	received his B. Eng. degree in Computer
	Engineering from University of Patras, Greece
	and his M.S. and Ph.D. degrees in Electrical
	Engineering from Florida Institute of Technology
	Melbourne, Florida. Dr. Plataniotis is currently
	a Professor with The Edward S. Rogers
	Sr. Department of Electrical and Computer
	Engineering at the University of Toronto in
	Toronto, Ontario, Canada, where he directs the
	Multimedia Laboratory. He holds the Bell Canada Endowed Chair in
	Multimedia since 2014. His research interests are primarily in the areas of
	image/signal processing, machine learning and adaptive learning systems,
	visual data analysis, multimedia and knowledge media, and effective
	computing. 
\end{IEEEbiography}

\end{document}